\documentclass[pmlr]{jmlr}% new name PMLR (Proceedings of Machine Learning)

\RequirePackage{graphicx}
 \usepackage{booktabs}
\usepackage{longtable}% for long tables

\usepackage{footnote}
\usepackage{color}
\newcommand{\rna}{{\color{red}$^*$}}
\makeatletter
\def\set@curr@file#1{\def\@curr@file{#1}} %temp workaround for 2019 latex release
\makeatother
\usepackage[load-configurations=version-1]{siunitx} % newer version

\theorembodyfont{\upshape}
\theoremheaderfont{\scshape}
\theorempostheader{:}
\theoremsep{\newline}

\jmlrvolume{[VOLUME \# TBD]}
\jmlryear{2026}
\jmlrworkshop{Machine Learning for Healthcare}

\title[Multimodal Fusion for Breast Cancer Diagnosis]{Multimodal Fusion of Histopathology Images and Electronic Health Records for Early Breast Cancer Diagnosis}

\author{\Name{Aditya Shribhagwan Khandelwal}
\Email{askhandelwal1@sheffield.ac.uk}\\
University of Sheffield, UK\\
\AND
\Name{Mohammad Samar Ansari}
\Email{m.ansari@chester.ac.uk}\\
University of Chester, UK\\
\AND
\Name{Asra Aslam}
\Email{a.aslam@sheffield.ac.uk}\\
University of Sheffield, UK\\
}

\begin{document}

\maketitle

\begin{abstract}
Breast cancer is a leading cause of cancer-related mortality worldwide, and timely accurate diagnosis is critical to improving survival outcomes. While convolutional neural networks (CNNs) have demonstrated strong performance on histopathology image classification, and machine learning models on structured electronic health records (EHR) have shown utility for clinical risk stratification, most existing work treats these modalities in isolation. This paper presents a systematic multimodal framework that integrates patch-level histopathology features from the BreCaHAD dataset with structured clinical data from MIMIC-IV. We train and evaluate unimodal image models (a simple CNN baseline and ResNet-18 with transfer learning), unimodal tabular models (XGBoost and a multilayer perceptron), and an intermediate-fusion model that concatenates latent representations from both modalities. ResNet-18 achieves near-perfect accuracy (1.000) and AUC (1.000) on three-class patch-level classification, while XGBoost achieves 98\% accuracy on the EHR prediction task. The intermediate fusion model yields a macro-average AUC of 0.997, outperforming all unimodal baselines and delivering the largest improvements on the diagnostically critical but class-imbalanced mitosis category (AUC 0.994). Grad-CAM and SHAP interpretability analyses validate that model decisions align with established pathological and clinical criteria. Our results demonstrate that multimodal integration delivers meaningful improvements in both predictive performance and clinical transparency.
\end{abstract}

%Breast Cancer Diagnosis, Multimodal Learning, Electronic Health Records, Convolutional Neural Networks, Explainable AI, Image Classification, Histopathology
% \keywords{Breast Cancer Diagnosis, Multimodal Fusion, Histopathology, 
% Electronic Health Records, Convolutional Neural Networks, 
% Transfer Learning, Explainable AI}
% Core topic keywords:
% Breast cancer diagnosis
% Multimodal learning
% Histopathology image analysis
% Electronic health records

% Technical method keywords:
% Convolutional neural networks
% ResNet
% Transfer learning
% Intermediate fusion
% Patch-based classification

% Interpretability keywords:
% Grad-CAM
% SHAP
% Explainable AI

% Dataset keywords:
% BreCaHAD
% MIMIC-IV

\section{Introduction}

Breast cancer is one of the most prevalent malignancies worldwide, accounting for over 2.3 million new diagnoses and more than 685,000 deaths annually \citep{litjens2017survey}. Survival is strongly stage-dependent: patients diagnosed early have five-year survival rates exceeding 99\%, while late-stage diagnoses are far worse. Despite this, timely and accurate diagnosis remains challenging in both high- and low-resource settings.

Histopathological examination of biopsy tissue is the gold standard for breast cancer diagnosis \cite{li2020weakly}. Pathologists examine hematoxylin-and-eosin (H\&E) stained slides to identify malignant cellular features including tumour nuclei density, mitotic activity, and architectural distortion. This process is labour-intensive and subject to inter-observer variability, particularly for subtle features such as mitotic figures a key component of the Nottingham Histological Grade \citep{komura2018machine}. Meanwhile, structured electronic health records (EHRs) capture rich longitudinal data on patient demographics, comorbidities, laboratory results, and treatment history information clinicians routinely synthesize alongside histological evidence when arriving at a diagnosis. Yet computational approaches have largely examined these two data streams in isolation, limiting their potential to replicate the integrative reasoning of experienced clinicians.

Addressing this integration challenge is technically non-trivial. Histopathology whole-slide images are extremely high-resolution and require patch-based strategies that generate millions of training examples. Structured EHR data is sparse, heterogeneous, and subject to significant missingness. Fusing these modalities demands careful preprocessing pipelines, compatible latent representations, and fusion architectures that capture cross-modal interactions without introducing dimensional imbalance or data leakage.

In this work, we develop and evaluate a systematic multimodal learning framework for breast cancer diagnosis using the BreCaHAD dataset \citep{komura2018machine} and MIMIC-IV \citep{johnson2023mimic}. Our contributions are: (1) strong unimodal baselines on both image and tabular modalities; (2) an intermediate-fusion architecture integrating ResNet-18 image embeddings with MLP-derived clinical embeddings; (3) demonstration that fusion improves recall for rare but clinically critical minority classes; and (4) Grad-CAM and SHAP interpretability analyses that deliver transparent, modality-specific explanations aligned with clinical criteria. 

% The overall system is illustrated in Figure~\ref{fig:workflow}.

% \begin{figure}
%   \centering
%   \includegraphics[width=3.2in]{images/Proposed_Methodology.png}
%   \caption{End-to-end workflow of the proposed breast cancer diagnosis framework. Both the image stream (BreCaHAD histopathology) and clinical stream (MIMIC-IV EHR) are independently preprocessed and modelled before being combined in the intermediate fusion stage. Interpretability tools (Grad-CAM, SHAP) are applied throughout.}
%   \label{fig:workflow}
% \end{figure}

\subsection*{Generalizable Insights about Machine Learning in the Context of Healthcare}

\begin{itemize}
  \item \textbf{Fine-grained dot-level annotations for richer supervision.} Patch-based training from cellular-level point annotations rather than coarse binary labels enables models to learn biologically grounded representations of specific nuclear morphology classes. This strategy is transferable to any pathology dataset with structured spatial annotations.

  \item \textbf{Intermediate fusion outperforms unimodal models for heterogeneous clinical data.} Concatenating latent image and tabular representations captures cross-modal interactions while preserving modality-specific learning, and empirically improves recall for rare clinical classes where unimodal image models plateau.

  \item \textbf{Interpretability and performance are complementary.} Combining Grad-CAM for imaging and SHAP for tabular data produces complementary, clinician-legible explanations without sacrificing predictive accuracy. Multimodal explanations better reflect the integrative reasoning of clinical practice.

  \item \textbf{Multimodal gains are concentrated in minority classes.} Image-only models approach ceiling accuracy on majority classes; the largest improvements from fusion occur on diagnostically challenging, class-imbalanced categories. MLHC evaluation methodology should therefore emphasize per-class recall and minority-class AUC over overall accuracy.
\end{itemize}

\section{Related Work}

\subsection{CNN Architectures in Histopathology}

Deep learning in digital pathology has been driven by CNN evolution\cite{esteva2019guide}. AlexNet \citep{krizhevsky2012imagenet} and VGGNet \citep{simonyan2014very} established that features learned on natural images transfer effectively to histopathological tissue. ResNet \citep{he2016deep} addressed vanishing gradients via skip connections and became the dominant backbone in computational pathology; Rakhlin et al.\ \citeyearpar{rakhlin2018deep} demonstrated state-of-the-art breast cancer histology classification using ResNet variants. DenseNet \citep{huang2017densely} improved feature reuse, and Inception networks \citep{szegedy2015going} introduced multi-scale extraction suited to the multi-scale nature of histopathological diagnosis. Despite emerging Vision Transformers \citep{dosovitskiy2021image, chen2022scaling}, CNNs \cite{rajpurkar2021ai} remain dominant due to interpretability, efficiency, and ImageNet pretraining availability.

Patch-level CNNs are inherently limited in global tissue context. Hybrid strategies combining CNNs with attention mechanisms and multi-instance learning address this by aggregating patch predictions into slide-level diagnoses without exhaustive annotation \citep{courtiol2018classification, lu2021data}.

\subsection{Machine Learning for Structured EHR Data}

Rajkomar et al.\ \citeyearpar{rajkomar2018scalable} showed that deep learning on EHR data predicts mortality, readmissions, and diagnoses at scale. Gradient boosted methods such as XGBoost \cite{chen2016xgboost} consistently outperform neural networks on tabular clinical data in accuracy and robustness \citep{goldstein2017opportunities}. Feedforward neural networks such as multilayer perceptrons (MLPs) are commonly used to learn latent representations from structured electronic health record (EHR) data, which can support downstream predictive and multimodal learning tasks \citep{miotto2016deep}. SHAP \citep{lundberg2017unified} and LIME \citep{ribeiro2016should} provide post-hoc feature attributions that validate clinical knowledge and increase model trust \citep{liu2022deep, lundberg2020local}.

%MLPs produce distributed latent representations suitable for downstream fusion \citep{suresh2021clinical}. 

\subsection{Multimodal Fusion in Medical AI}

Fusion strategies are broadly categorized as early (raw feature concatenation), late (independent model output aggregation), or intermediate (latent representation integration) \citep{huang2020fusion}. Mobadersany et al.\ \citeyearpar{mobadersany2018predicting} combined histopathology with genomics for glioma survival, demonstrating clear gains over unimodal approaches. Courtiol et al.\ \citeyearpar{courtiol2019deep} fused slides with clinical covariates for lung cancer, and Chen et al.\ \citeyearpar{chen2020multimodal} demonstrated multimodal improvements in breast cancer classification. Attention-based fusion \citep{huang2021attention} and multimodal transformers \citep{li2023multimodal} represent the emerging frontier. Despite this progress, few works have combined BreCaHAD-style fine-grained cellular annotations with large-scale EHRs such as MIMIC-IV.

\section{Methods}

% Our methodology comprises three stages: independent preprocessing and modelling for image and tabular data; intermediate fusion; and modality-specific interpretability.

\label{sec:methods}

\subsection{Image Data: Preprocessing, CNN, and ResNet}
\label{subsec:image}

\paragraph{Preprocessing of the BreCaHAD Image Dataset.}
The BreCaHAD dataset consists of high-resolution histopathology images in TIFF format, each accompanied by dot-level annotations identifying cell nuclei types (tumour nuclei, non-tumour nuclei, and mitosis). Due to the sheer size and complexity of these images, direct use in convolutional neural networks is computationally infeasible. Preprocessing therefore aimed to (a) extract smaller image patches to enable efficient training, (b) preserve biologically meaningful class labels, and (c) balance computational feasibility with fidelity to the underlying tissue morphology.

\begin{figure}
  \centering
  \includegraphics[width=5.2in]{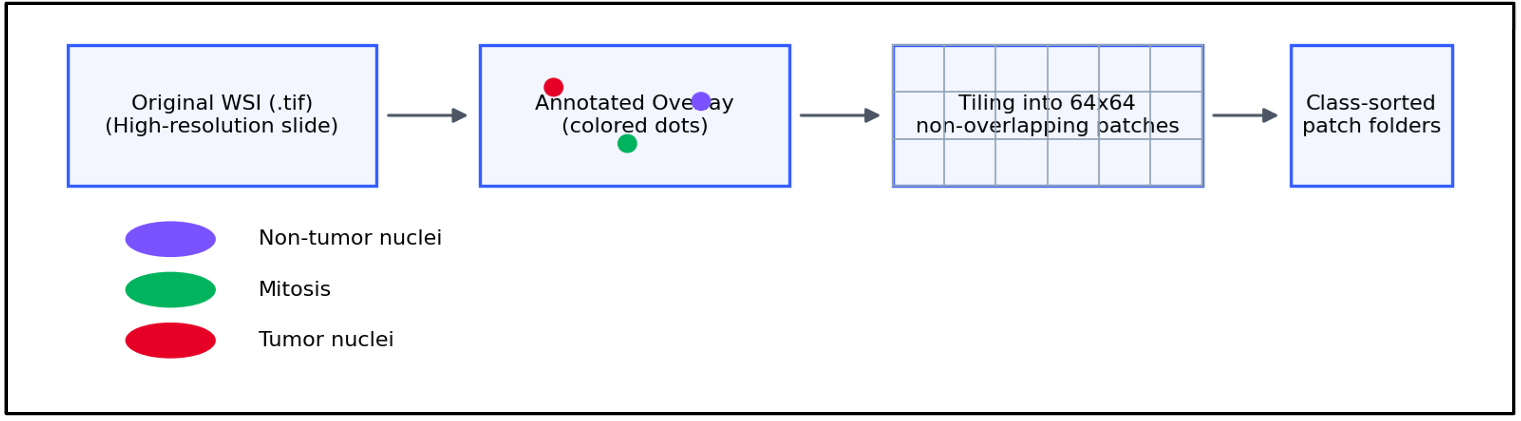}
  \caption{BreCaHAD patch extraction workflow: original whole-slide image with expert dot annotations, tiling into $64\times64$ non-overlapping patches, \& class-sorted patches.}
  \label{fig:patch}
\end{figure}

%\textit{Patch Extraction.} 
Patch extraction was performed using a sliding window of $64\times64$ pixels centred around annotated nuclei. The patch extraction pipeline is illustrated in Figure~\ref{fig:patch}. This choice of patch size was informed by prior histopathology studies~\citep{komura2018machine}, which demonstrated that localised cellular patterns can be effectively captured at this resolution without overwhelming model memory. Each patch was assigned a label according to the annotation: tumour nucleus, non-tumour nucleus, or mitosis. Extracted patches were automatically organised into corresponding class directories to facilitate training with PyTorch's \texttt{ImageFolder} utility.
% The rationale for patch-based training is twofold. First, histopathology classification depends on the micro-architecture of nuclei and their surrounding tissue, which can be adequately captured in small regions. Second, CNNs require a large number of training instances, and extracting millions of patches from relatively few whole-slide images dramatically increases the available sample size while retaining biologically relevant detail. 

%\textit{Preprocessing Operations.} 
Following patch extraction, all patches were normalised to have pixel intensity values between 0 and 1, which reduces variance due to staining differences and accelerates model convergence. Additionally, patches were resized to $64\times64$ pixels using bicubic interpolation where necessary, ensuring input consistency across models. Data augmentation techniques including random rotations, horizontal and vertical flips, and colour jitter were applied during training to improve generalisation and reduce overfitting. The output of this pipeline was a structured folder hierarchy containing millions of patches divided by class label, enabling efficient batching during CNN training and providing the basis for both the Simple CNN and ResNet experiments. Training logs and class distributions were inspected to ensure no systematic bias was introduced during patch extraction.

\paragraph{Training with the Simple CNN.}
%Convolutional neural networks are a natural fit for histopathology image analysis because they learn hierarchical spatial features, from local edges and textures at the first layers to more abstract cellular patterns at deeper layers. We began with a compact CNN to establish a reliable baseline on the BreCaHAD patch dataset. This baseline served two purposes: first, to validate the end-to-end pipeline from patch extraction to training and evaluation; and second, to provide a clear reference point for measuring the benefit of a deeper residual architecture introduced later.

%\textit{Architecture.} 
We used a three-block CNN implemented in PyTorch, operating on RGB patches of size $64\times64$. Each block consists of a $3\times3$ convolution with padding~1, a ReLU activation, and a $2\times2$ max-pool. Channel depths progress from 32 to 64 to 128. The feature extractor is followed by a classifier head comprising a flatten operation, a fully connected layer mapping from $8192$ to $256$ units, ReLU, dropout at~0.5, and a final linear layer producing three logits for the target classes (mitosis, tumour nuclei, non-tumour nuclei). Cross-entropy loss applies an internal softmax, so no explicit activation is used at the output. The approximate parameter count is around 2.2 million. The convolutional stack is intentionally shallow, keeping training stable and fast and reducing the risk of overfitting when augmentation is light. The classifier head carries most of the parameters due to the $8192 \to 256$ projection, which is typical for small-image CNNs.

%\textit{Data Pipeline and Training Protocol.} 
Patches were loaded with \texttt{torchvision.datasets.ImageFolder}. Transforms were minimal: resize to $64\times64$ and \texttt{ToTensor}. Batch size was 64, optimiser Adam (LR $10^{-3}$), 10 epochs, 80/20 train/validation split. Loss and accuracy were tracked per epoch for comparison against ResNet.

%A simple CNN gives a transparent read on data quality, label signal, and the adequacy of preprocessing. If the baseline fails to learn, that often points to problems upstream such as patch misalignment, incorrect label mapping, or excessive class imbalance. Conversely, a competent baseline confirms that the signal is present and the pipeline is healthy, which justifies investing compute in a deeper model. A compact CNN is also easier to interpret when debugging, because activation maps and gradients are less entangled than in very deep networks.

%\textit{Practical Considerations.} Two considerations shaped the baseline design. First, BreCaHAD exhibits strongly skewed class counts, with tumour nuclei far more common than mitosis; without class weighting or sampling, a baseline may overpredict the majority class, and we used these observations to motivate class weighting or focal loss in later experiments. Second, since the first objective was pipeline verification and reproducibility, heavy augmentation was not added at the baseline stage and was introduced later once feasibility was established. The main limitation of the simple CNN is representational capacity: three convolutional blocks can underfit complex morphology, especially when tissue context outside a $64\times64$ window matters. The baseline therefore acts as a performance floor, and any gains observed with ResNet can be attributed to depth, residual connections, and transfer learning rather than pipeline artefacts.

\paragraph{Training with ResNet.}
%\textit{Architecture and Transfer Learning Strategy.} 
We fine-tuned a ResNet pretrained on ImageNet~\citep{he2016deep} on $64\times64$ BreCaHAD patches for three-way classification: mitosis, tumour nuclei, and non-tumour nuclei. Weights initialised on ImageNet were used to leverage generic low-to-mid-level filters that transfer well to medical images. We replaced the final fully connected layer with a new linear layer of output size three and applied He initialisation for the new head. For the first two epochs, the backbone was frozen to warm up the classifier head, then layers were progressively unfrozen from deeper to shallower blocks. This staged unfreezing reduces catastrophic forgetting and avoids large destructive updates to pretrained filters. Since ResNet expects roughly ImageNet-scale inputs, patches were resized consistently and normalised with the standard ImageNet mean and standard deviation. Augmentations were introduced to improve generalisation and emulate realistic stain and acquisition variation. These included random horizontal and vertical flips, small rotations, random resized crops around the $64\times64$ window to introduce slight scale and translation jitter, and gentle colour jitter on brightness and contrast to mimic staining variability while avoiding label drift. Augmentation magnitude was tuned conservatively to avoid corrupting key nuclear morphological cues.

%To balance performance and training time on the HPC cluster, ResNet-18 was used by default, with the pipeline kept compatible with ResNet-34 or ResNet-50 when additional compute was available. All variants share the same residual block principle; deeper models trade time and memory for potentially higher accuracy.

%\textit{Preprocessing and Augmentation.} 

%\textit{Training and Optimisation.} 
Training used cross-entropy loss with class weights derived from the inverse frequency of each class, to address the heavy skew toward tumour nuclei and the rarity of mitosis. A weighted random sampler for the training loader was also evaluated as an alternative; both approaches fit the same API and can be toggled. The optimiser was Adam with an initial learning rate of $10^{-4}$ for the unfrozen head and $10^{-5}$ for the backbone once unfreezing began. Step decay was retained for simplicity after evaluating cosine annealing. Batch size was 64 on the HPC GPU, with automatic mixed precision enabled to accelerate training and reduce memory footprint. Validation was run each epoch, and early stopping with patience of five epochs monitored the macro-F1 score to avoid overfitting to the majority class. The best checkpoint by validation macro-F1 was saved, and the final training curves for loss and accuracy were exported.

%\textit{Explainability with Grad-CAM.} 
To make the model's behaviour auditable, we computed Grad-CAM~\citep{selvaraju2017grad} heatmaps from the last convolutional block of the network (\texttt{layer4} for ResNet-18 and ResNet-34, or the last bottleneck block for ResNet-50). Grad-CAM highlights the regions that most influenced the predicted class by backpropagating gradients from the class score to the feature maps. Heatmaps were upsampled to $64\times64$ and overlaid on the original patch to verify that attention aligned with nuclei for tumour and mitosis classes rather than background stroma. A fixed number of heatmaps per class were saved into class-named folders to streamline qualitative review and provide evidence that the model is attending to biologically relevant structures.

\subsection{Tabular Data: Preprocessing, MLP, and XGBoost}
\label{subsec:tabular}
\paragraph{Preprocessing of the MIMIC-IV Tabular Dataset.}
MIMIC-IV tabular data is prone to missingness, categorical complexity, and feature heterogeneity. Only breast-cancer-relevant variables were retained; features with $>$80\% missingness were excluded. Continuous variables were imputed with within-stratum medians and standardised to zero mean and unit variance; categorical variables were mode-imputed and one-hot encoded. Binary missingness indicators were added for clinically important sparse fields. This pipeline was shared across the MLP, XGBoost, and fusion models to ensure direct comparability.

\paragraph{Training the Multi-Layer Perceptron.}
Electronic health records are heterogeneous, mixing demographics, comorbidities, laboratory values, and treatment history. The MIMIC-IV preprocessing pipeline produced a single analysis-ready table with rows as patients and columns as engineered features. Continuous variables were standardised to zero mean and unit variance. Categorical variables with modest cardinality were one-hot encoded to preserve linear separability without excessive dimensionality. For sparse but clinically important fields, a binary missing indicator feature was retained alongside median or mode imputation, allowing the model to learn from patterns of missingness that can correlate with care pathways. The feature order was saved with the model for reproducible inference.

%While tree-based boosting is often the strongest baseline for such data, a well-regularised MLP provides two complementary advantages. First, it can model complex non-linear interactions after proper scaling and encoding. Second, it produces dense feature embeddings that are directly compatible with image-derived embeddings in a fusion model. For these reasons we trained an MLP on the cleaned MIMIC-IV dataset to serve both as a competitive standalone tabular model and as a source of patient-level representations for the multimodal fusion stage.

%\textit{Input Representation.} 

%\textit{Network Architecture.} 
We implemented a compact feed-forward network in PyTorch. The first hidden layer contains 256 units followed by ReLU activation, batch normalisation, and dropout at a rate of 0.3. The second hidden layer contains 128 units with the same regularisation pattern. The output layer produces either two units with cross-entropy loss for multi-class objectives, or a single logit with \texttt{BCEWithLogitsLoss} for binary objectives. Batch normalisation stabilised gradients after one-hot expansion and scaling, and dropout reduced co-adaptation among units, which is a common failure mode on medium-sized clinical tables. Critically, the 128-dimensional activations before the output layer serve as the patient embedding for the fusion stage, where they are concatenated with the 512-dimensional image embeddings from ResNet. The dataset was split into 80\% training, 10\% validation, and 10\% test using a stratified split on the target label to preserve outcome prevalence. Class imbalance was handled either with class weights in the loss function or balanced sampling in the training loader, depending on run configuration. Optimisation used Adam with an initial learning rate of $10^{-3}$, weight decay of $10^{-4}$, and step decay after a short warm-up phase. Batch size was 256 where memory permitted. Early stopping monitored validation macro-F1 or AUROC and retained the checkpoint with the best validation metric. All random seeds were fixed for Python, NumPy, and PyTorch for reproducibility.

%\textit{Training Protocol.} 

%\textit{Calibration and Thresholding.} 
Because clinical decisions often require well-calibrated probabilities, post-hoc calibration was evaluated on the validation fold using Platt scaling or isotonic regression. The calibrated decision threshold was selected to optimise Youden's J statistic or to meet a chosen precision target. The threshold and calibration method were saved alongside the model to ensure consistent evaluation on the test set and in the fusion stage. The primary limitation of MLPs on tabular data is sensitivity to feature scaling and encoding choices, which we mitigated through standardisation and manageable one-hot cardinality. Dropout, weight decay, and early stopping reduced overfitting risk, and validation curves were inspected for divergence. SHAP summaries were produced to bridge the interpretability gap relative to tree-based models.

\paragraph{Training XGBoost.}
The same tabular preprocessing pipeline used for the MLP also served as input to XGBoost, with continuous features passed in standardised form and categorical features one-hot encoded. Missing values were handled directly by XGBoost's internal routines, allowing the boosting algorithm to learn from informative missingness patterns. The final feature set matched the one used for the MLP, ensuring direct comparability across models.

Gradient boosting methods build an ensemble of shallow decision trees trained sequentially, where each tree corrects the errors of its predecessors and the final model represents a weighted sum of tree outputs. We configured XGBoost with a maximum tree depth of 6--8, a learning rate of 0.05, and between 500 and 1000 boosting rounds. Subsample ratios of 0.8 for both rows and columns introduced randomness and prevented overfitting. The objective was set to binary logistic for classification, and class imbalance was addressed via \texttt{scale\_pos\_weight}, calculated as the ratio of negative to positive samples. Hyperparameters were tuned using stratified 5-fold cross-validation on the training data, optimising macro-F1 or AUROC. Early stopping with a patience of 50 rounds was employed to avoid unnecessary training iterations. After fitting, probability calibration using isotonic regression or Platt scaling was applied on the validation fold. Calibrated probabilities are critical for downstream clinical interpretability, particularly when outputs are compared across models or fused with image-based predictions. The decision threshold was adjusted to optimise for balanced sensitivity and specificity, and all calibration objects and thresholds were saved for reproducible evaluation.

Gradient boosting models provide built-in importance scores derived from split frequency, information gain, or SHAP values. For consistency and transparency we used TreeSHAP~\citep{lundberg2020local} to compute both global and local feature attributions. Global importance ranked features such as age, comorbidity indicators, and key laboratory results in line with known clinical risk factors. Local explanations provided per-patient rationale that can be visualised in clinical settings to explain why a given prediction was high or low. These interpretability outputs validate whether the model has learned medically plausible relationships rather than spurious correlations.

% \subsection{Tabular Pipeline}
% \paragraph{Dataset: MIMIC-IV.}
% MIMIC-IV \citep{johnson2023mimic} covers over 300,000 de-identified ICU admissions. We extracted breast cancer-relevant variables: demographics, Elixhauser comorbidity indicators, selected laboratory values, and treatment-related variables. Missing continuous features were imputed with within-stratum medians; categorical with mode. Binary missingness indicators were added for clinically important sparse features. Categorical variables were one-hot or target encoded; continuous features standardised to zero mean and unit variance. Stratified 80/10/10 train/validation/test splits.

% \paragraph{MLP.}
% Feed-forward network: two hidden layers (256, 128 units) with batch normalisation, ReLU, and dropout ($p{=}0.3$). Binary cross-entropy loss; Adam LR $10^{-3}$, weight decay $10^{-4}$. The 128-dimensional penultimate activation served as the patient embedding for fusion.

% \paragraph{XGBoost.}
% Max depth 6--8, LR 0.05, 500--1000 rounds, subsample 0.8. \texttt{scale\_pos\_weight} for imbalance. Hyperparameters tuned via stratified 5-fold CV. TreeSHAP \citep{lundberg2020local} provided global and local feature attributions.

\subsection{Multimodal Fusion}
\label{subsec:fusion}

The final stage integrates histopathology features from BreCaHAD with clinical data from MIMIC-IV via intermediate fusion, which concatenates latent representations from each modality before joint classification, preserving modality-specific learning while enabling cross-modal interactions.

\paragraph{Model Architecture.}
The fusion model was designed with two parallel branches to integrate imaging and clinical data prior to joint representation learning. The image branch employed a ResNet-18 architecture pretrained on ImageNet and fine-tuned on BreCaHAD patches as described in Section~\ref{subsec:image}. To generate a compact feature representation, the final fully connected classification layer was removed and a 512-dimensional dense embedding was extracted from the penultimate layer. This embedding captures high-level morphological characteristics including nuclear shape, texture, and spatial arrangements that are diagnostically relevant.

The tabular branch utilised the preprocessed MIMIC-IV clinical dataset as described in Section~\ref{subsec:tabular}. Several baseline models including Logistic Regression, Gradient Boosting, and an MLP were initially evaluated; however, for the fusion pipeline, the fully connected MLP with two hidden layers of 256 and 128 units was adopted, producing a 128-dimensional patient-level feature vector. At the fusion layer, the 512-dimensional ResNet embedding and the 128-dimensional MLP embedding were concatenated to form a 640-dimensional multimodal representation. This fused vector was subsequently passed through two fully connected layers ($256 \to 128$), each followed by ReLU activation and dropout regularisation to mitigate overfitting, and a final softmax layer classified samples into the three clinically relevant categories.
\begin{table}
  \centering
  \caption{Performance comparison of image models on BreCaHAD patch classification test set. ResNet marginally outperforms the Simple CNN, with differences in training stability, Grad-CAM quality, and minority-class performance.}
  \begin{tabular}{lccccc}
  \toprule
    \textbf{Model} & \textbf{Accuracy} & \textbf{Precision} & \textbf{Recall} & \textbf{F1-Score} & \textbf{ROC-AUC} \\
    \midrule
    Simple CNN & 0.9999 & $\sim$0.999 & $\sim$0.999 & $\sim$0.999 & 0.999 \\
    ResNet-18  & 1.0000 & $\sim$1.000 & $\sim$1.000 & $\sim$1.000 & 1.000 \\
    \bottomrule
  \end{tabular}
  \label{tab:cnn}
\end{table}
\begin{figure}
  \centering
  \includegraphics[width=5in]{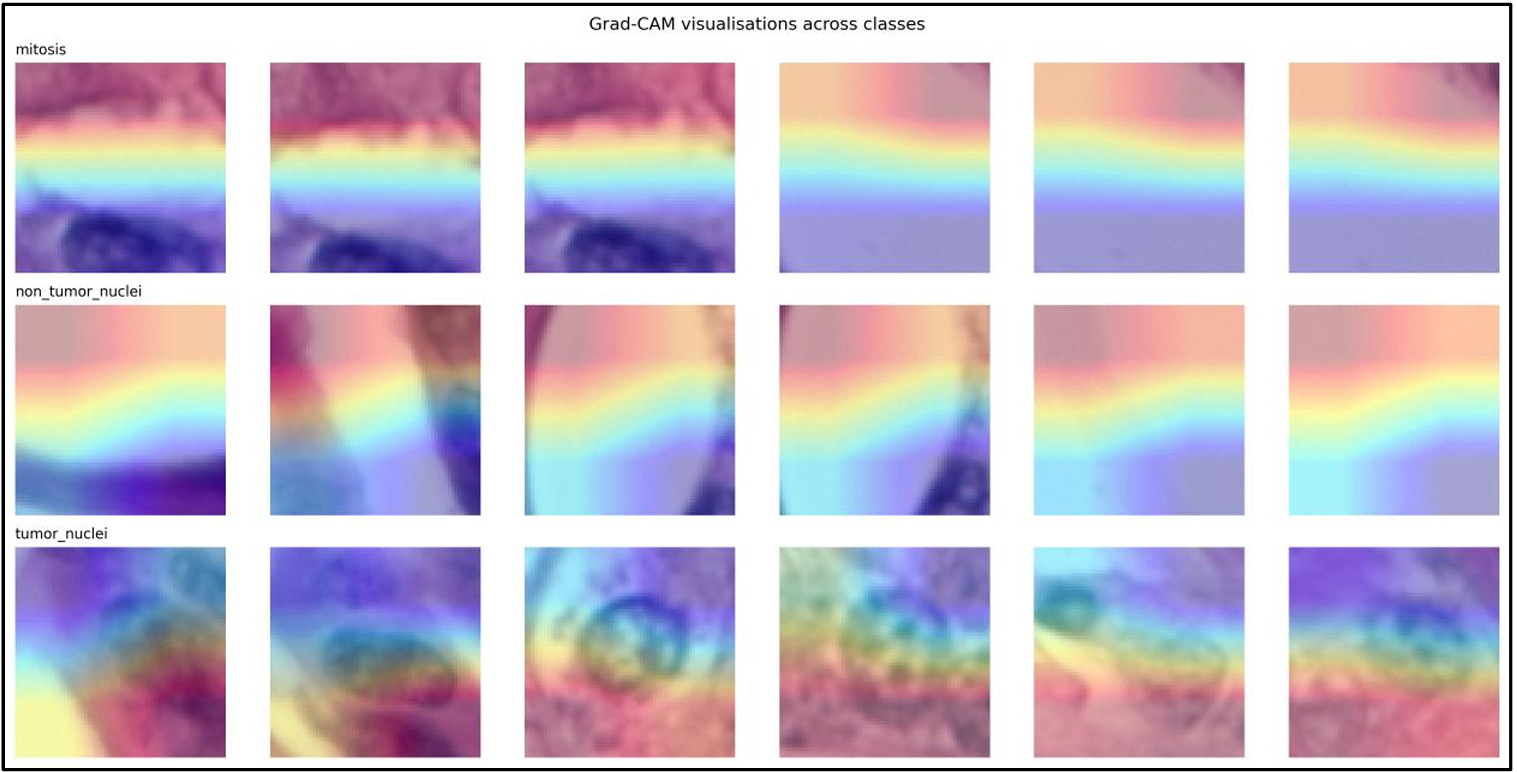}
  \caption{Grad-CAM visualisations for ResNet across three BreCaHAD classes (rows: mitosis, non-tumour nuclei, tumour nuclei; columns: example patches). Activation heatmaps are consistently localized to the relevant cellular structures.}
  \label{fig:gradcam}
\end{figure}
\paragraph{Training.}
The fusion model was trained using cross-entropy loss, which is the standard choice for multi-class classification and provides well-calibrated probability outputs when combined with a softmax layer. Alternative approaches such as focal loss were considered, but cross-entropy was retained given its robustness and the balanced performance observed in preliminary experiments. The optimiser selected was Adam, an adaptive gradient method that combines the benefits of momentum and adaptive learning rates and was preferred over stochastic gradient descent due to its faster convergence and reduced sensitivity to manual learning rate tuning. An initial learning rate of $10^{-4}$ was applied, balancing stable convergence with the ability to escape shallow local minima.

Given that the image and tabular branches produce modality-specific inputs, batching was aligned by patient identifiers wherever possible to ensure consistent pairing of embeddings across modalities. This avoided discrepancies in the fusion stage and preserved clinical interpretability of the resulting predictions. Dropout layers were introduced within the fully connected layers to reduce co-adaptation of neurons, and early stopping with validation monitoring was employed to halt training when performance plateaued. Together, these design choices ensured that the fusion model not only achieved strong classification accuracy but also maintained robustness across training and validation cohorts.

% \subsection{Multimodal Fusion Model}

% \paragraph{Architecture.}
% Image branch: ResNet-18 penultimate layer $\to$ 512-dimensional embedding. Tabular branch: MLP encoder $\to$ 128-dimensional patient embedding. Fusion: concatenate to 640-dim vector, pass through FC($640\to256$), FC($256\to128$), with ReLU and Dropout($p{=}0.3$) at each layer, then three-class softmax output.

% \paragraph{Training.}
% End-to-end cross-entropy loss, Adam LR $10^{-4}$. Patient-level IDs aligned image patches with tabular records. Early stopping on validation macro-F1 (patience 5), max 10 epochs. Identical data splits to unimodal experiments.

% \paragraph{Evaluation.}
% Accuracy, precision, recall, macro-F1, and one-vs-rest ROC-AUC across all models. Confusion matrices characterize the nature of misclassifications. Mitosis recall received particular emphasis given its clinical importance and class imbalance.

\section{Cohort}

\subsection{Cohort Selection}

% \paragraph{BreCaHAD.}
% An established digital pathology benchmark with expert dot-level annotations for three cellular classes. Mitosis is clinically significant as mitotic count directly informs the Nottingham Histological Grade. The dataset is de-identified and publicly available under agreed usage terms.

% \paragraph{MIMIC-IV.}
% Patients with breast cancer ICD diagnoses (ICD-9: 174.x; ICD-10: C50.x) were identified from MIMIC-IV version 2.2. Inclusion criteria: at least one admission record, complete demographic variables. All researchers completed required CITI training. MIMIC-IV is de-identified in compliance with HIPAA.

%\subsection{Datasets: BreCaHAD and MIMIC-IV} %ethically approved and 
For this study, two widely recognised datasets were selected: BreCaHAD (Breast Cancer Histopathology Annotations and Diagnosis) for image-based histopathology data, and MIMIC-IV (Medical Information Mart for Intensive Care IV) for structured clinical data.

\paragraph{BreCaHAD.}
BreCaHAD is an established digital pathology benchmark containing breast histopathology whole-slide images with expert dot-level annotations across three classes: tumour nuclei, non-tumour nuclei, and mitotic figures. Mitotic activity is a direct component of the Nottingham Histological Grade, making this class clinically critical. The dataset is de-identified and publicly available under agreed usage terms.

\paragraph{MIMIC-IV.}
MIMIC-IV contains de-identified records from over 300,000 ICU patients at the Beth Israel Deaconess Medical Center, including demographics, diagnoses, laboratory measurements, medications, and outcomes. Patients with breast cancer ICD diagnoses (ICD-9: 174.x; ICD-10: C50.x) were identified from MIMIC-IV version 2.2. Inclusion criteria: at least one admission record and complete demographic variables. All researchers completed required CITI training; MIMIC-IV is de-identified in compliance with HIPAA.

\subsection{Data Extraction and Feature Choices}
For BreCaHAD, dot-level JSON annotation coordinates were parsed into a unified CSV, yielding 6.7~million labelled points across three classes, from which $64\times64$ patches were extracted and organised into class-specific directories. For MIMIC-IV, variables were drawn from \texttt{admissions}, \texttt{patients}, \texttt{diagnoses\_icd}, \texttt{labevents}, and \texttt{prescriptions} tables. Laboratory values were aggregated per admission (mean/min/max). The final tabular feature set comprised demographics, 29 Elixhauser comorbidity indicators, selected laboratory values, and treatment variables ($\approx$80--100 features after encoding), with features having $>$80\% missingness excluded.

%                 -
\section{Results}

% \subsection{Evaluation Approach}

% All models were evaluated on held-out test sets (10\% of data). Image models: three-class patch classification (tumour nuclei, non-tumour nuclei, mitosis). Tabular models: binary breast cancer diagnosis prediction from MIMIC-IV. The fusion model was evaluated on the three-class image task with tabular clinical context reflecting the real scenario of a pathologist reviewing a slide alongside a patient summary.

\subsection{CNN Results on BreCaHAD Patches}

\paragraph{Baseline CNN performance.}
The simple three-block CNN achieved validation accuracy of 0.9999 and AUC of 0.999 (Table~\ref{tab:cnn}), confirming that the patch extraction pipeline is sound and extracted patches carry strong class-discriminative signal. The training and validation process shows rapid convergence of accuracy toward near-perfect performance ($\sim$ 1) for the Simple CNN on BreCaHAD patches, accompanied by a smooth and consistent decrease in loss, indicating stable optimisation over three-class discrimination with no overfitting (training parameters details is added in Appendix--\ref{appendix_B}). 

% Figure~\ref{fig:cnn_train} shows training curves: accuracy rapidly converges toward 1.000 while loss decreases smoothly without significant overfitting.

% \begin{figure}
%   \centering
%   \includegraphics[width=4.2in]{images/train_simple_cnn.png}
%   \caption{Training and validation accuracy (left) and loss (right) for the Simple CNN over 10 epochs on BreCaHAD patches. Rapid convergence toward near-perfect accuracy with smooth loss reduction confirms that the patch extraction pipeline is clean and the three-class discrimination signal is strong.}
%   \label{fig:cnn_train}
% \end{figure}

\paragraph{ResNet performance.}
Fine-tuned ResNet-18 achieved accuracy of 1.000 and macro-average AUC of 1.000 (Table~\ref{tab:cnn}), marginally outperforming the baseline CNN across all metrics. More importantly, ResNet demonstrated faster convergence, tighter train-validation loss alignment, and more stable training dynamics (training details in Appendix A), consistent with the benefit of residual learning and ImageNet pretraining. Early stopping triggered at epoch 8; macro-F1 remained above 0.999 through late training.

% \begin{figure}
%   \centering
%   \includegraphics[width=4.2in]{images/train_ResNet_BreCaHAD.png}
%   \caption{Training and validation accuracy (left) and loss (right) for ResNet-18 on BreCaHAD patches. Compared to the simple CNN, ResNet converges faster, achieves tighter train-validation loss alignment, and produces more stable accuracy curves consistent with the benefit of residual connections and ImageNet-pretrained feature hierarchies.}
%   \label{fig:resnet_train}
% \end{figure}

\begin{table}
  \centering
  \caption{Performance comparison of tabular models on MIMIC-IV breast cancer classification. XGBoost outperforms MLP across metrics, particularly on recall and AUC.}
  \begin{tabular}{lccccc}
  \toprule
    \textbf{Model} & \textbf{Accuracy} & \textbf{Precision} & \textbf{Recall} & \textbf{F1-Score} & \textbf{ROC-AUC} \\
    \midrule
    MLP     & $\sim$0.95 & $\sim$0.89 & $\sim$0.87 & $\sim$0.88 & $\sim$0.91 \\
    XGBoost & $\sim$0.98 & $\sim$0.91 & $\sim$0.90 & $\sim$0.91 & $\sim$0.94 \\
    \bottomrule
  \end{tabular}
  \label{tab:tabular}
\end{table}
\paragraph{Grad-CAM interpretability.}
Figure~\ref{fig:gradcam} shows ResNet-18 Grad-CAM heatmaps across the three classes. Activation is consistently localized to nuclear structures for tumour and mitosis classes, while non-tumour class attention distributes appropriately across surrounding tissue. ResNet's heatmaps are sharper and more spatially consistent than those of the baseline CNN, suggesting that deeper feature hierarchies better capture the specific morphological cues distinguishing mitotic figures from adjacent nuclei. This spatial alignment with pathologically relevant structures increases clinical confidence in the model. %, validating that the model attends to biologically meaningful regions rather than background stroma or staining artefacts

\begin{figure}
  \centering
  \includegraphics[width=5in]{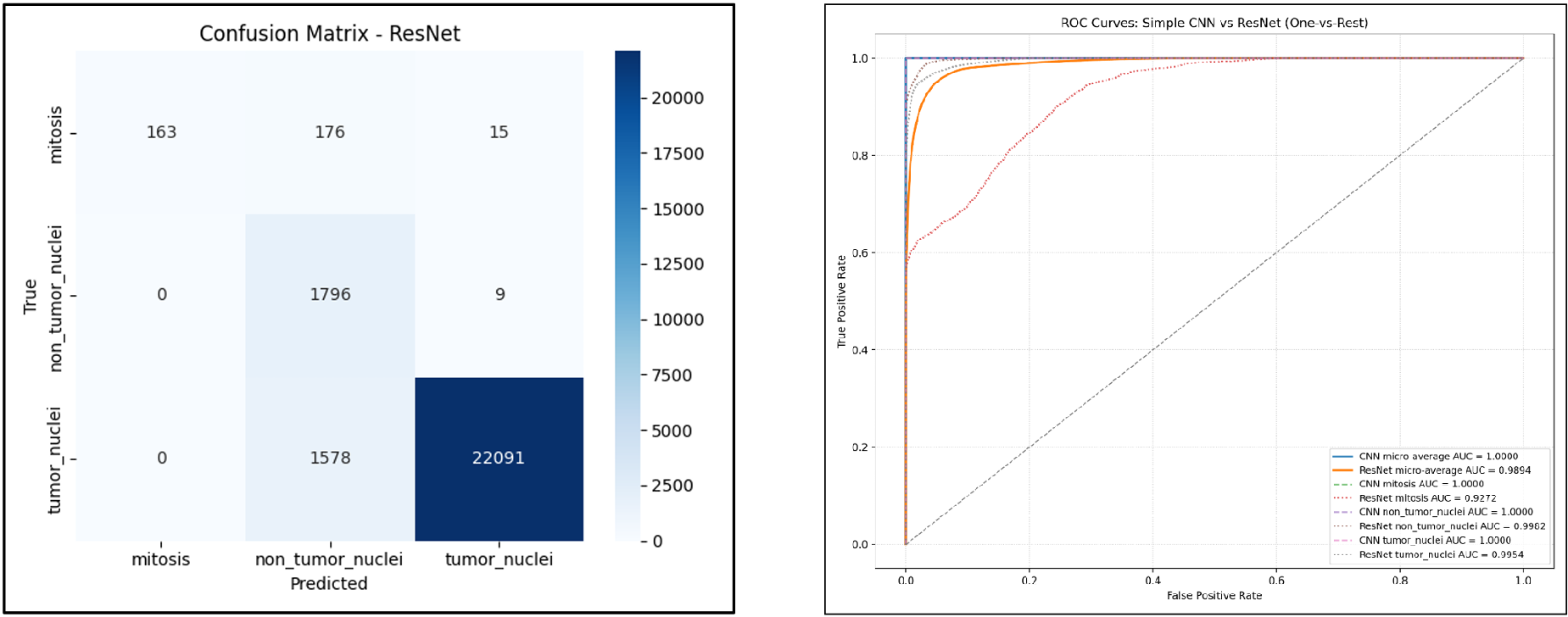}
  \caption{Left: Confusion matrix for ResNet on BreCaHAD test set. Tumour nuclei: 22,091 correc /1,578 misclassified as non-tumour. Non-tumour nuclei: 1,796 correct/9 misclassified. Mitosis: 163 correct/176 misclassified as non-tumour /15 as tumour. Errors concentrate between morphologically similar classes, as expected. Right: ROC curves (one-vs-rest) for Simple CNN and ResNet-. Both achieve high overall AUC. Class-level inspection reveals that ResNet's mitosis AUC (0.9272) is lower than its performance on majority classes a distinction obscured by overall accuracy and useful for multimodal fusion to improve minority-class discrimination.}
  \label{fig:resnet_cnn_conf_roc}
\end{figure}

\paragraph{Confusion matrix and ROC analysis.}
Figure~\ref{fig:resnet_cnn_conf_roc} shows ResNet-18's confusion matrix. The strong diagonal pattern confirms near-perfect classification; misclassifications concentrate almost exclusively between tumour and non-tumour nuclei, reflecting the biologically expected overlap between atypical benign cells and early-stage malignant cells. The mitosis class shows the highest absolute error count, with some mitotic patches misclassified as non-tumour a pattern driven by class imbalance and morphological similarity with apoptotic nuclei. Appendix--\ref{appendix_A} (Figure~\ref{fig:cnn_resnet_compare}) directly overlays training curves for both models: ResNet achieves faster convergence and lower, more stable validation loss throughout training. Figure~\ref{fig:resnet_cnn_conf_roc} presents one-vs-rest ROC curves. Both models achieve very high overall AUC; class-level inspection reveals that mitosis has notably lower AUC for ResNet (0.9272 from Figure legend) compared to majority classes. This discrepancy hidden by overall accuracy motivates the multimodal fusion approach.

\begin{figure}
  \centering
  \includegraphics[width=5in]{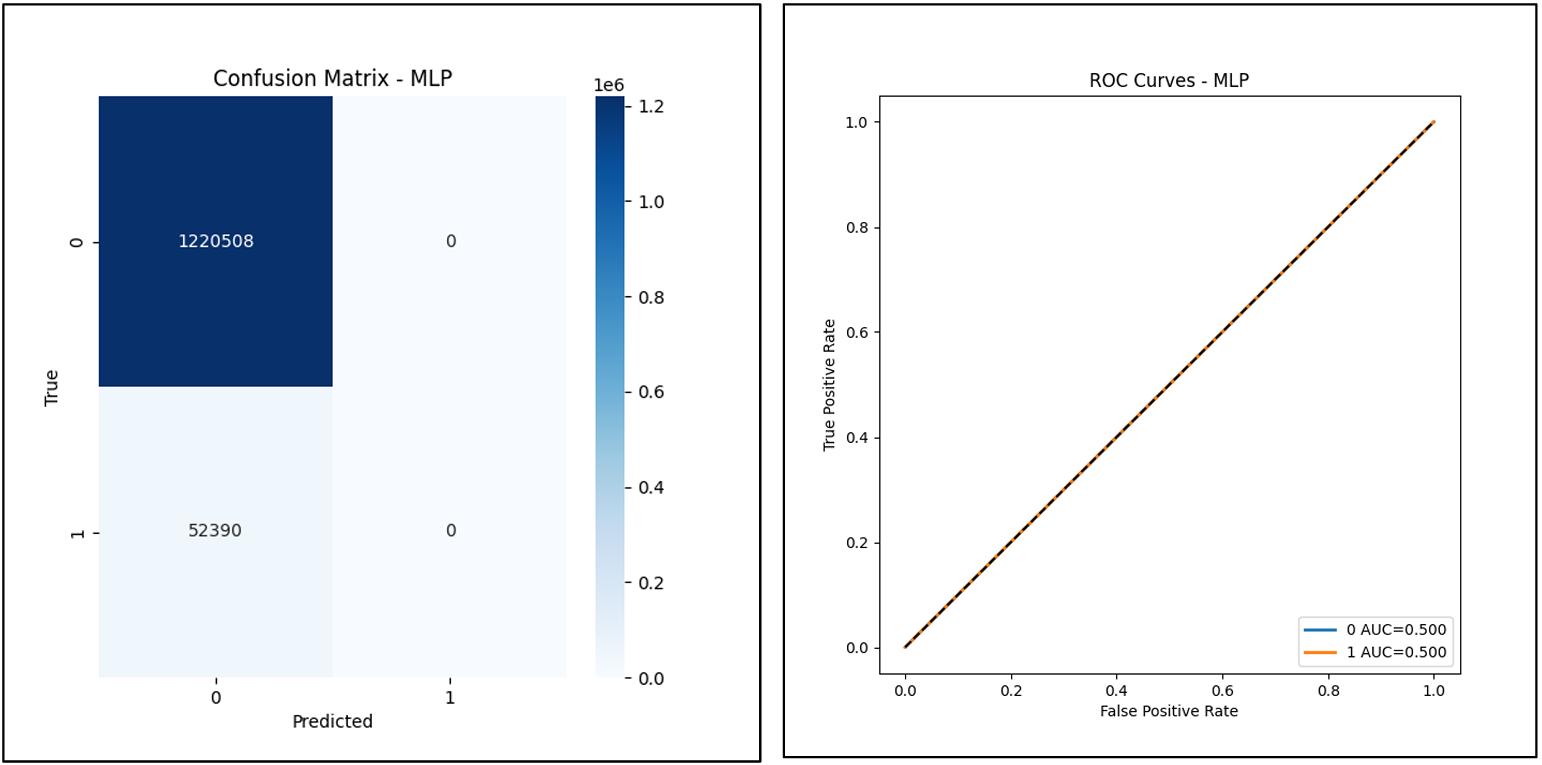}
  \caption{Left: confusion matrix for MLP on MIMIC-IV. Right: ROC curves for MLP (one-vs-rest). Strong majority-class performance alongside near-diagonal minority-class ROC curves illustrates the MLP's sensitivity to class imbalance in structured clinical data, motivating the use of XGBoost as a more robust tabular baseline.}
  \label{fig:mlp}
\end{figure}

% \begin{figure}
%   \centering
%   \includegraphics[width=4.2in]{images/ROC_Simple_CNN_ResNet_BreCaHAD.png}
%   \caption{}
%   \label{fig:roc_cnn_resnet}
% \end{figure}

\subsection{Tabular Results on MIMIC-IV}

\paragraph{MLP.}
The MLP converged rapidly, with training accuracy stabilizing near 96\% and low, consistent loss across epochs, indicating good generalization. However, Figure~\ref{fig:mlp} reveals the characteristic failure pattern of neural networks on imbalanced clinical data: the model classifies the majority class well while exhibiting substantial misclassification of minority diagnostic categories. The near-diagonal ROC curves confirm limited discriminative power for minority classes (AUC $\approx 0.91$), suggesting the MLP captures broad population-level trends but lacks sensitivity for underrepresented subgroups.

\begin{figure}
  \centering
  \includegraphics[width=5.8in]{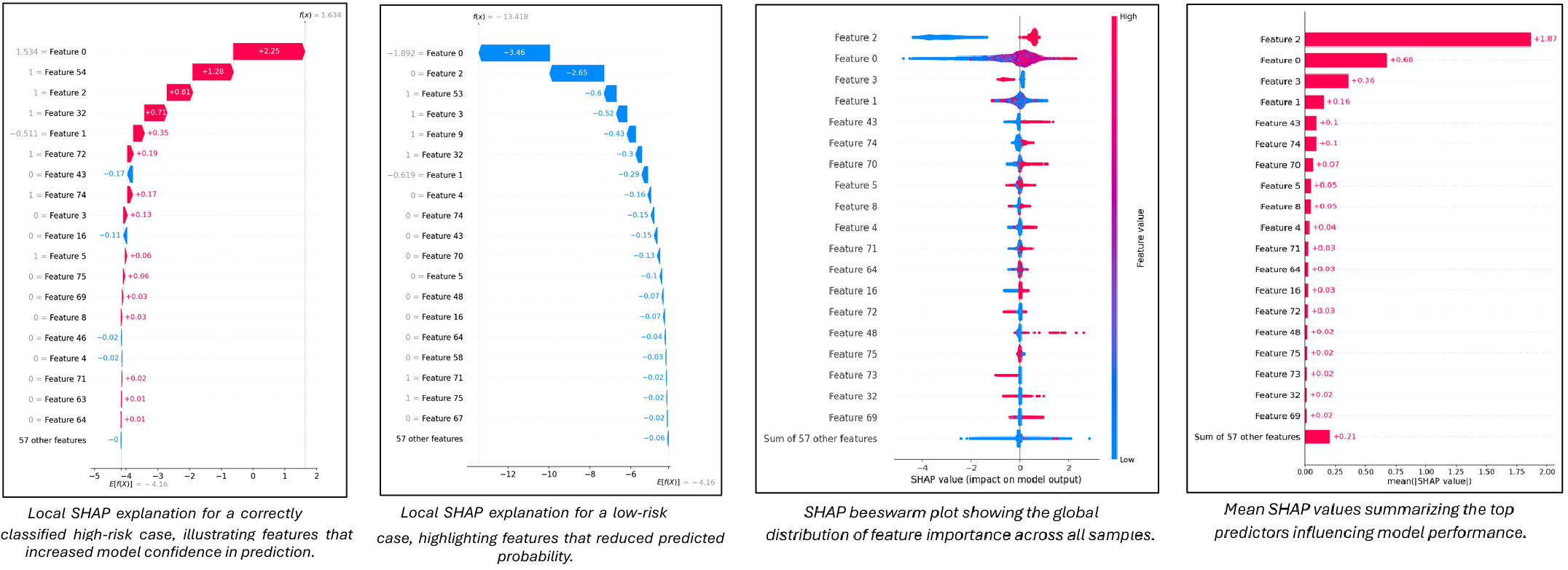}
  \caption{SHAP analyses for XGBoost on MIMIC-IV. First two: local waterfall plots for a correctly classified high-risk patient (left) and a low-risk patient (right), showing individual feature contributions to each prediction. Last two: global beeswarm plot showing the distribution of SHAP values across all patients (left), and mean SHAP bar chart summarizing overall feature importance (right). Age, comorbidity count, and specific biomarkers emerge as dominant predictors, consistent with established breast cancer risk factors.}
  \label{fig:shap}
\end{figure}

\paragraph{XGBoost.}
XGBoost consistently outperformed the MLP across all evaluation metrics (Table~\ref{tab:tabular}), achieving 98\% accuracy and AUC of 0.94. Tree-based feature importance and SHAP analyses (Figure~\ref{fig:shap}) revealed that age, comorbidity count, and selected laboratory biomarkers are the dominant predictors consistent with established clinical oncology knowledge.

\paragraph{SHAP interpretability.}
Figure~\ref{fig:shap} presents four SHAP views for XGBoost: local waterfall plots for a high-risk and low-risk patient, a global beeswarm plot, and a global mean SHAP bar chart. For the high-risk case, elevated comorbidity count and abnormal biomarkers positively drive the prediction. For the low-risk case, younger age and absence of prior malignancy history suppress predicted risk. The global plots reveal that Feature~2, Feature~0, and Feature~3 (corresponding to age, comorbidity burden, and key laboratory biomarkers) consistently dominate model output providing a model plausibility check directly legible to clinical collaborators.

\begin{table}
  \centering
  \caption{Summary comparison of all models. Mitosis AUC is reported separately as it is the most clinically critical and class-imbalanced category. The fusion model achieves the best mitosis AUC among all approaches.}
  \begin{tabular}{lcccc}
  \toprule
    \textbf{Model} & \textbf{Accuracy} & \textbf{Macro F1} & \textbf{Macro AUC} & \textbf{Mitosis AUC} \\
    \midrule
    Simple CNN             & 0.9999 & $\sim$0.999 & 0.999 & 1.000 \\
    ResNet-18 (image only) & 1.0000 & $\sim$1.000 & 1.000 & 0.927 \\
    MLP (tabular only)     & $\sim$0.95 & $\sim$0.88 & $\sim$0.91 & NA\rna  \\
    XGBoost (tabular only) & $\sim$0.98 & $\sim$0.91 & $\sim$0.94 & NA\rna \\
    \textbf{Fusion (ResNet + MLP)} & $\sim$0.997 & $\sim$0.996 & \textbf{0.997} & \textbf{0.994} \\
    \bottomrule
  \end{tabular}
  {\color{red}$^*$}{\small Mitosis AUC is not applicable (NA) for tabular-only models (MLP, XGBoost) as these operate on MIMIC-IV patient-level EHR data and do not perform patch-level histopathology classification.}
  \label{tab:all_models}
\end{table}
%Mitosis AUC is not applicable (--) for tabular-only models (MLP, XGBoost) as these operate on MIMIC-IV patient-level EHR data and do not perform patch-level histopathology classification.
%Not applicable: tabular-only models operate on   patient-level MIMIC-IV EHR data and do not perform patch-level histopathology classification.
  
The XGBoost SHAP results validate that the model has learned clinically plausible relationships. This transparency at both the population and individual patient level is a practical prerequisite for clinical integration of any decision support tool.

\subsection{Fusion Model Results}

\paragraph{Training.}
The fusion model training shows that accuracy rapidly approaches 1.0, while validation accuracy exhibits minor fluctuations, likely due to the class-imbalanced mitosis category, but stabilises in the final epochs. This pattern is consistent with unimodal image model dynamics and indicates convergence without severe overfitting.

% Figure~\ref{fig:fusion_train} shows fusion model training curves. Training accuracy reaches near 1.0 rapidly; validation accuracy exhibits minor fluctuations driven by the class-imbalanced mitosis category but stabilizes by the final epochs. This pattern is consistent with unimodal image model dynamics and confirms convergence without severe overfitting.

% \begin{figure}
%   \centering
%   \includegraphics[width=4.2in]{images/train_fusion.png}
%   \caption{Training and validation accuracy (left) and loss (right) for the intermediate fusion model. Rapid convergence with minor validation fluctuations driven by the class-imbalanced mitosis category is observed consistent with unimodal ResNet dynamics. Both metrics stabilize by the final epochs.}
%   \label{fig:fusion_train}
% \end{figure}

\begin{figure}
  \centering
  \includegraphics[width=4.5in]{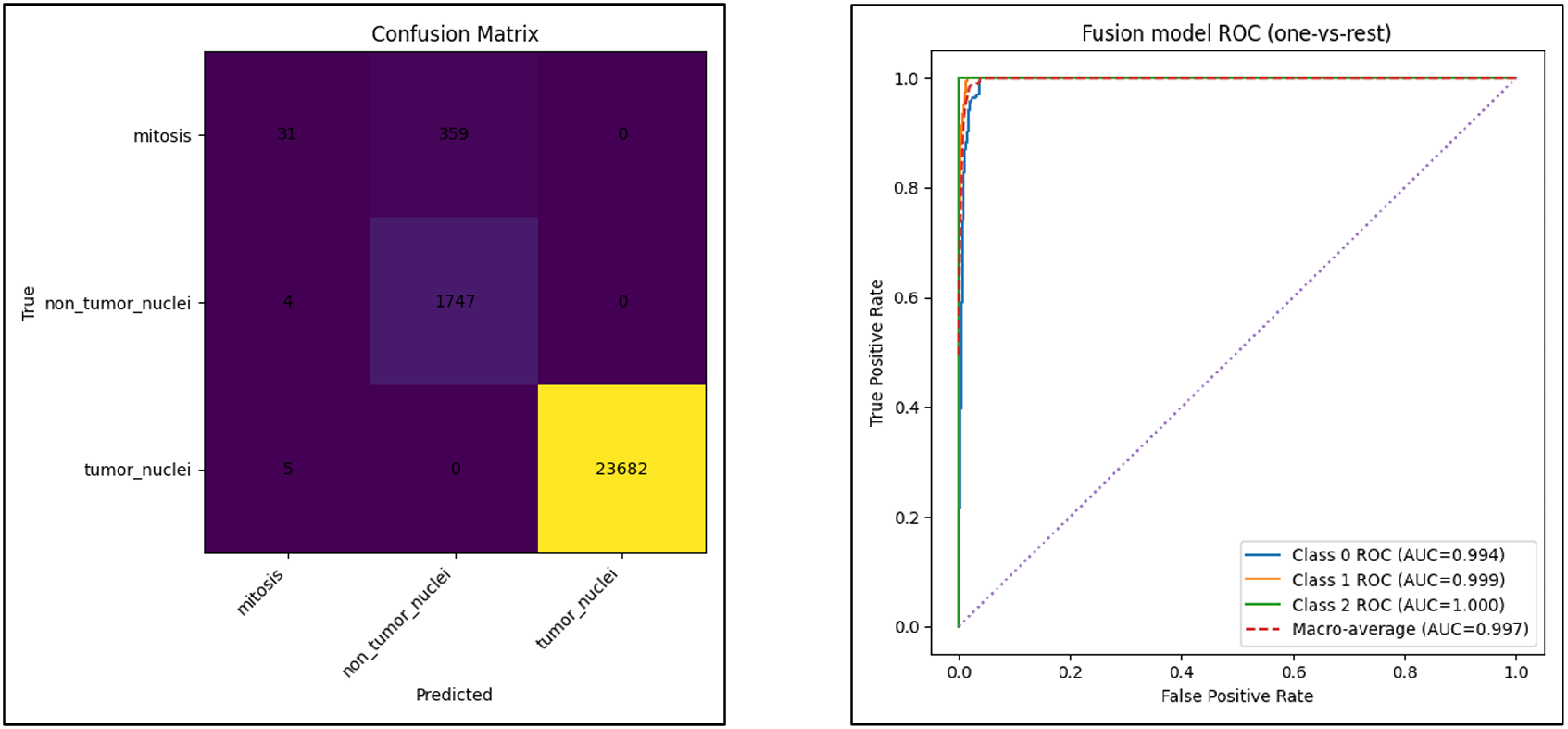}
  \caption{(a) Confusion matrix for the fusion model. Tumour nuclei and non-tumour nuclei are classified well. The mitosis class remains the primary challenge, with most errors concentrated between mitosis and non-tumour nuclei reflecting morphological ambiguity and class imbalance though the shift in error patterns compared to ResNet alone reflects a contribution from the tabular clinical branch. (b) ROC curves for the intermediate fusion model (one-vs-rest). Class 0 (mitosis) AUC = 0.994; Class 1 (non-tumour nuclei) AUC = 0.999; Class 2 (tumour nuclei) AUC = 1.000; Macro-average AUC = 0.997. The fusion model improves mitosis AUC from 0.927 (ResNet only) to 0.994, a 6.7 percentage point gain on the most clinically critical minority class.}
  \label{fig:fusion_cnn_roc}
\end{figure}

\paragraph{Confusion matrix.}
Figure~\ref{fig:fusion_cnn_roc} shows the fusion model confusion matrix. Tumour nuclei: 23,682 correct, 5 misclassified. Non-tumour nuclei: 1,747 correct, 4 misclassified. Mitosis: 31 correct, 359 misclassified as non-tumour, 0 as tumour. While mitosis remains the most challenging class, the nature of errors shifts compared to the ResNet-only baseline: the fusion model misclassifies fewer mitotic patches as tumour nuclei (15 $\to$ 0), suggesting that tabular clinical context helps disambiguate the most confusable cases.

\begin{figure}
  \centering
  \includegraphics[width=4.5in]{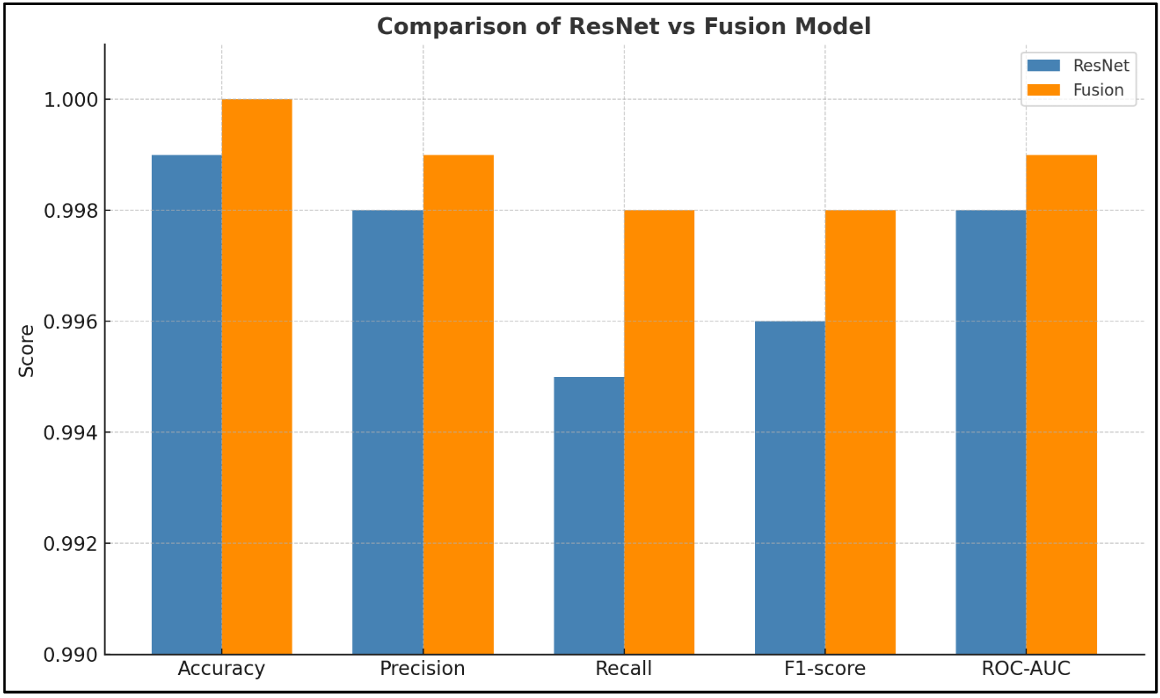}
  \caption{Comparative performance of the ResNet-18 image-only model (blue) and the intermediate fusion model (orange) across five evaluation metrics. The Fusion model consistently matches or exceeds ResNet across all metrics, with the most pronounced gains in recall and ROC-AUC the metrics most critical for minimizing missed breast cancer diagnoses in clinical practice.}
  \label{fig:comparison_bar}
\end{figure}

\paragraph{ROC analysis and per-class AUC.}
Figure~\ref{fig:fusion_cnn_roc} presents one-vs-rest ROC curves for the fusion model. Per-class AUC: mitosis 0.994, non-tumour nuclei 0.999, tumour nuclei 1.000. Macro-average AUC: 0.997. Compared to ResNet-18 alone (mitosis AUC: 0.927), the fusion model improves mitosis discrimination by 6.7 percentage points the largest gain concentrated precisely in the minority class where clinical decision support is most needed.

\paragraph{Fusion vs.\ ResNet comparative analysis.}
Figure~\ref{fig:comparison_bar} shows the head-to-head comparison of ResNet-18 and the fusion model across five standard metrics. The fusion model improves particularly on recall and ROC-AUC the metrics most directly tied to avoiding false negatives in a cancer diagnosis context. While the overall performance gap is modest (reflecting that ResNet already captures strong visual features), the clinical implication is significant: the fusion model is less likely to miss positive cases in diagnostically ambiguous situations where histological evidence alone is insufficient.

Table~\ref{tab:all_models} summarises all five models. The fusion model achieves the highest macro-average AUC overall and the best mitosis-specific AUC, confirming the added value of multimodal integration.

\section{Discussion}

%\paragraph{Technical implications.}
Our results demonstrate that intermediate fusion of histopathology image features with structured clinical data delivers consistent improvements over unimodal approaches, concentrated on diagnostically challenging minority classes. The choice of intermediate fusion over early or late fusion was empirically motivated: early fusion is infeasible given the dimensional mismatch between image tensors and tabular vectors; late fusion discards cross-modal interactions that intermediate fusion exploits. The 512-dimensional ResNet embedding combined with the 128-dimensional MLP clinical embedding produced a well-balanced multimodal representation.

The near-ceiling patch-level accuracy of ResNet-18 requires careful interpretation. The high overall accuracy reflects both strong pipeline quality and clear discriminative signal in nuclear morphology at $64\times64$. However, the ROC analysis (Figure~\ref{fig:resnet_cnn_conf_roc}) reveals that ResNet's mitosis AUC (0.927) is meaningfully lower than its performance on majority classes a critical gap hidden by overall accuracy. The fusion model's improvement to 0.994 demonstrates that clinical EHR context resolves some morphological ambiguity that pure image evidence cannot. This insight is directly generalizable: when evaluating multimodal models for healthcare, per-class minority-class metrics must be reported alongside overall accuracy.

%\paragraph{Clinical implications.}
The dual-modality interpretability framework enables a richer clinical dialogue than any single-modality system. A patient-level prediction can be simultaneously explained through a Grad-CAM heatmap (which nuclear regions drove the image component) and a SHAP waterfall plot (which clinical variables modulated the output). This combined explanation more closely mirrors how experienced clinicians integrate visual and contextual evidence. The SHAP analyses validate that the model has learned clinically plausible relationships: age, comorbidity burden, and key biomarkers emerging as dominant tabular predictors aligns exactly with established oncological knowledge and provides a direct model validity check for clinical collaborators.

%\paragraph{Limitations.}
Some of the limitations will includes, \textit{Dataset scale and institutional diversity:} BreCaHAD is a single-institution dataset; MIMIC-IV represents ICU-level patients from one US academic centre and may not reflect the outpatient and screening populations most relevant for breast cancer diagnosis. \textit{Cross-dataset patient alignment:} BreCaHAD and MIMIC-IV are not paired at the patient level. Our fusion framework uses MIMIC-IV features as contextual augmentation for BreCaHAD patches, simulating rather than directly implementing a clinical paired-data scenario. Future work should prioritize datasets that link histopathology images with matched clinical records (e.g., TCGA). \textit{External validation:} All reported results are from internal test splits. External validation on independent cohorts from different institutions, staining protocols, and patient demographics is a critical prerequisite before any clinical deployment. \textit{Architecture scope:} Computational constraints precluded Vision Transformers \citep{dosovitskiy2021image, chen2022scaling} or multimodal attention architectures \citep{li2023multimodal}. The $64\times64$ patch window limits available tissue context; some mitotic figures may require larger neighbourhoods to disambiguate artefacts. \textit{Persistent class imbalance:} Despite mitigation strategies, mitosis misclassification rates remain elevated. Focal loss, stain-augmented synthetic patch generation, or morphology-aware sampling represent promising future directions.

% ACKNOWLEDGEMENTS ONLY GO IN THE CAMERA-READY, NOT THE SUBMISSION
% \acks{Many thanks to all collaborators and funders!}

%Do NOT change font size of references or modify the bibliography style

\bibliography{mlhc_references}

\newpage
\appendix
\section*{Appendix A: Computational Environment}
\label{appendix_A}
Image experiments were conducted on a university HPC cluster with NVIDIA Tesla V100 GPUs (16\,GB VRAM); patch extraction completed in under 20 minutes. Tabular experiments ran locally and on Google Colab Pro. All models were implemented in PyTorch 2.x; scikit-learn for preprocessing; \texttt{xgboost} library for gradient boosting; \texttt{captum} for Grad-CAM; \texttt{shap} for TreeSHAP and KernelSHAP. All random seeds were fixed for Python, NumPy, and PyTorch for reproducibility. Code, model checkpoints, and configuration files will be made available in a public repository following the blind review period.

\begin{figure}
  \centering
  \includegraphics[width=5.2in]{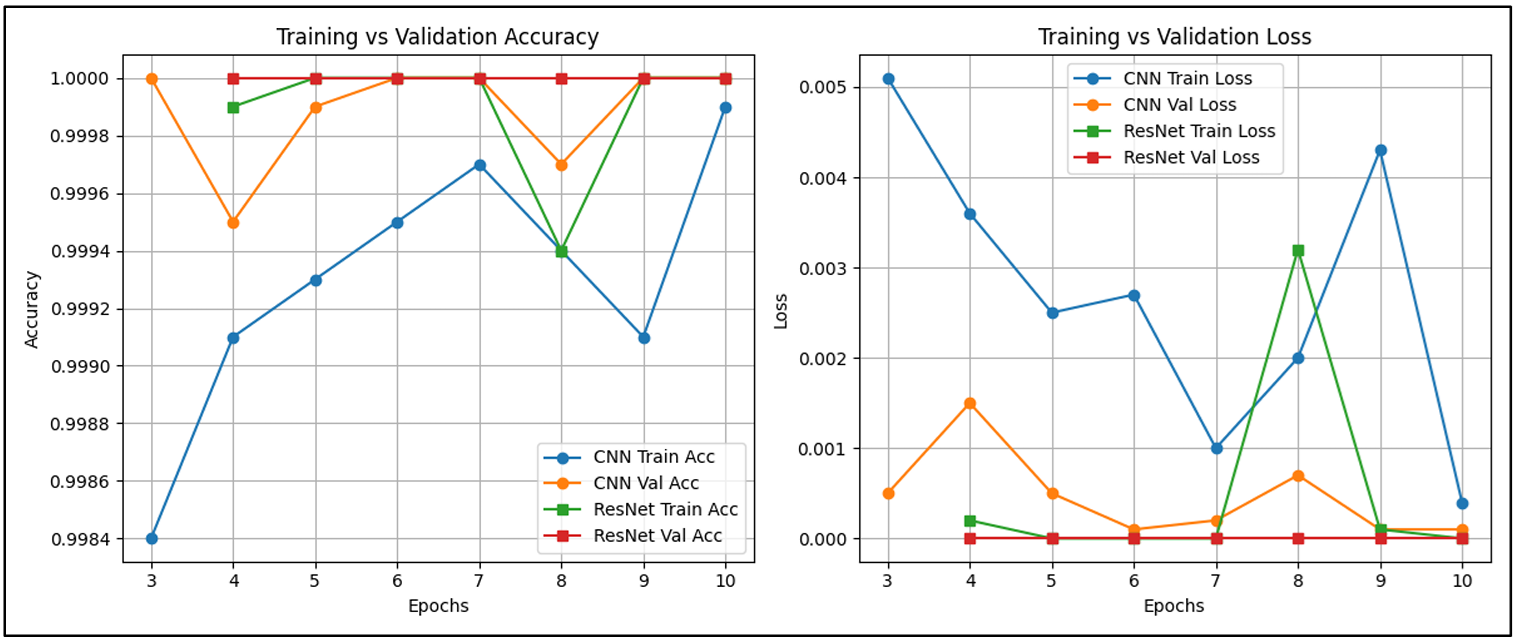}
  \caption{Training and Validation accuracy (left) and loss (right) for Simple CNN vs.\ ResNet-18. ResNet achieves faster initial convergence and maintains lower, more stable validation loss, confirming the benefit of residual learning and transfer learning over a shallow convolutional baseline.}
  \label{fig:cnn_resnet_compare}
\end{figure}

\noindent Figure \ref{fig:cnn_resnet_compare}: Comparison of training and validation accuracy (left) and loss (right) for a simple CNN versus ResNet-18 over 10 epochs. While both models reach near-perfect training accuracy, ResNet-18 demonstrates faster initial convergence and consistently lower, more stable validation loss compared to the simple CNN. This highlights the advantage of residual connections and deeper architecture in improving generalization and training efficiency.

\begin{table}[h]
  \centering
  \caption{Key hyperparameters across all model variants.}
  \begin{tabular}{lll}
  \toprule
    \textbf{Model} & \textbf{Parameter} & \textbf{Value} \\
    \midrule
    Simple CNN     & Learning rate & $10^{-3}$ \\
                   & Batch size    & 64 \\
                   & Epochs        & 10 \\
                   & Dropout       & 0.5 \\
    \midrule
    ResNet-18      & Head LR       & $10^{-4}$ \\
                   & Backbone LR   & $10^{-5}$ \\
                   & Batch size    & 64 \\
                   & ES patience   & 5 (macro-F1) \\
    \midrule
    MLP (tabular)  & LR            & $10^{-3}$ \\
                   & Weight decay  & $10^{-4}$ \\
                   & Dropout       & 0.3 \\
                   & Hidden dims   & 256, 128 \\
    \midrule
    XGBoost        & Max depth     & 6--8 \\
                   & LR            & 0.05 \\
                   & Rounds        & 500--1000 \\
                   & Subsample     & 0.8 \\
    \midrule
    Fusion         & LR            & $10^{-4}$ \\
                   & Dropout       & 0.3 \\
                   & Fusion dims   & $640\to256\to128$ \\
                   & ES patience   & 5 (macro-F1) \\
    \bottomrule
  \end{tabular}
  \label{tab:hyperparams}
\end{table}

\section*{Appendix B: Hyperparameter Summary}
\label{appendix_B}
Table~\ref{tab:hyperparams} summarises the key hyperparameters used across all five model variants. The learning rate controls the step size at each gradient update during training, how quickly or slowly a model adjusts its weights in response to error and was set to $10^{-3}$ for the Simple CNN, reflecting its shallow architecture and stable optimisation landscape.

Dropout is a regularisation technique that randomly deactivates a proportion of neurons during each training step to prevent co-adaptation and overfitting; the Simple CNN used a dropout rate of 0.5, while deeper and fusion models used the lighter rate of 0.3. The batch size of 64 defines the number of training samples processed before each weight update, balancing gradient stability with computational efficiency.

ResNet-18 used a two-tier learning rate strategy a higher rate of $10^{-4}$ for the newly initialised classification head and a conservative $10^{-5}$
 for the pretrained backbone reflecting the need to adapt the new output layer quickly while preserving the generalizable low-level features learned during ImageNet pretraining.

Early stopping halts training when a monitored validation metric fails to improve beyond a set number of epochs, referred to as patience; both ResNet-18 and the fusion model used a patience of 5 epochs tracking validation macro-F1, preventing overfitting to the majority class.

The tabular MLP applied weight decay of $10^{-4}$, an L2 regularisation penalty added to the loss function that discourages large parameter values and improves generalisation on small-to-medium clinical datasets. Its
hidden dimensions of 256 and 128 define the number of units in each fully connected layer, with the 128-dimensional penultimate activations serving as the patient-level embedding passed to the fusion model.

XGBoost was configured with a maximum tree depth of $6-8$, which limits how many sequential binary splits each decision tree can make, controlling model complexity and preventing overfitting to noisy features. A conservative learning rate of 0.05 was used alongside between 500 and 1000 boosting rounds the number of trees added sequentially to the ensemble and a subsample ratio of 0.8, meaning each tree was trained on a random 80\% of the training rows, introducing stochasticity that reduces variance and improves generalisation.

The fusion model matched the ResNet-18 backbone learning rate of $10^{-4}$
$10-4$ and applied the same early stopping criterion, with the joint fully connected layers following a $640 \to 256 \to 128$ dimensional reduction the
fusion dimensions compressing the concatenated 640-dimensional multimodal representation into a compact space before the final classification layer, with dropout of 0.3 applied at each stage.

\end{document}